%% file: main.tex
\pdfoutput=1

\documentclass[10pt,twocolumn,letterpaper]{article}

\usepackage[pagenumbers]{cvpr} 

\usepackage{graphicx}
\usepackage{amsmath}
\usepackage{amssymb}
\usepackage{booktabs}

\usepackage{times}
\usepackage{epsfig}
\usepackage{comment}
\usepackage{color}
\usepackage{xcolor}
\usepackage{caption}
\usepackage{subcaption}

\makeatletter
\@namedef{ver@everyshi.sty}{}
\makeatother
\usepackage{tikz}
\usetikzlibrary{tikzmark}

\usepackage[pagebackref,breaklinks,colorlinks]{hyperref}

\usepackage[capitalize]{cleveref}
\crefname{section}{Sec.}{Secs.}
\Crefname{section}{Section}{Sections}
\Crefname{table}{Table}{Tables}
\crefname{table}{Tab.}{Tabs.}

\newcommand{\myparagraph}[1]{\vspace{0pt} \noindent \textbf{#1}}

\def\cvprPaperID{9902} 
\def\confName{CVPR}
\def\confYear{2023}

\begin{document}


\title{
BlendGAN: Learning and Blending the Internal Distributions of Single Images by Spatial Image-Identity Conditioning
}

\author{Idan Kligvasser\\
Verily Life Sciences\\
\and
Tamar Rott Shaham\\
Technion, MIT\
\and
Noa Alkobi\\
Technion\\
\and
Tomer Michaeli\\
Technion\\
}

\maketitle

\input{sections/abstract}
\input{sections/intro}

\input{sections/related}
\input{sections/method}
\input{sections/experiments}
\input{sections/conclusion}


{\small
\bibliographystyle{ieee_fullname}
\bibliography{bib}
}

\end{document}

%% file: sections/abstract.tex
\input{figures/figure1/figure1_v4}

\begin{abstract}
Training a generative model on a single image has drawn significant attention in recent years. Single image generative methods are designed to learn the internal patch distribution of a single natural image at multiple scales. These models can be used for drawing diverse samples that semantically resemble the training image, as well as for
solving many image editing and restoration tasks that involve that particular image. Here, we introduce an extended framework, which allows to simultaneously learn the internal distributions of several images, by using a single model with spatially varying image-identity conditioning. Our BlendGAN opens the door to applications that are not supported by single-image models, including morphing, melding, and structure-texture fusion between two or more arbitrary images. 
\end{abstract}

%% file: figures/figure1/figure1_v4.tex
\begin{figure*}[ht]
	\centering
	\captionsetup[subfigure]{labelformat=empty,justification=centering,aboveskip=1pt,belowskip=1pt}
	\begin{subfigure}[t]{0.15\textwidth}
		\centering
		\caption{Training image id \#1}
		\fbox{\includegraphics[width=1\linewidth,]{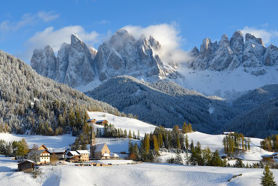}}
	\end{subfigure}
	\begin{subfigure}[t]{0.69\textwidth}
	\centering
	\caption{Image morphing}
	\begin{subfigure}[t]{0.22\textwidth}
		\centering
		\includegraphics[width=1\linewidth,]{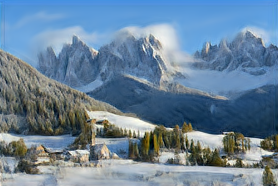}
	\end{subfigure}
	\begin{subfigure}[t]{0.22\textwidth}
		\centering
		\includegraphics[width=1\linewidth,]{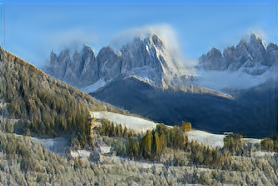}
	\end{subfigure}
	\begin{subfigure}[t]{0.22\textwidth}
		\centering
		\includegraphics[width=1\linewidth,]{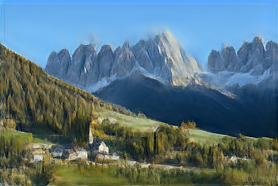}
	\end{subfigure}
	\begin{subfigure}[t]{0.22\textwidth}
		\centering
		\includegraphics[width=1\linewidth,]{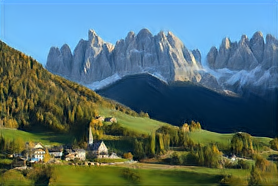}
	\end{subfigure}
	\end{subfigure}
	\begin{subfigure}[t]{0.15\textwidth}
		\centering
		\caption{Training image id \#2}
		\fbox{\includegraphics[width=1\linewidth,]{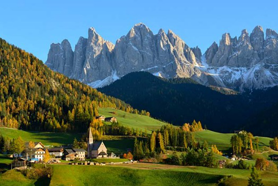}}
	\end{subfigure}
	\par\smallskip
	\begin{subfigure}[t]{0.15\textwidth}
		\centering
		\caption{Sample from id \#1}
		\includegraphics[width=1\linewidth,]{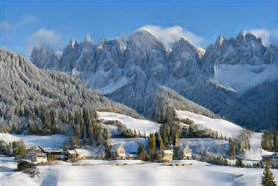}
	\end{subfigure}
	\begin{subfigure}[t]{0.69\textwidth}
	\centering
	\caption{Sample morphing}
	\begin{subfigure}[t]{0.22\textwidth}
		\centering
		\includegraphics[width=1\linewidth,]{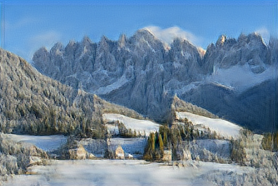}
	\end{subfigure}
	\begin{subfigure}[t]{0.22\textwidth}
		\centering
		\includegraphics[width=1\linewidth,]{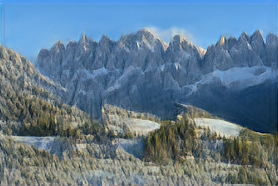}
	\end{subfigure}
	\begin{subfigure}[t]{0.22\textwidth}
		\centering
		\includegraphics[width=1\linewidth,]{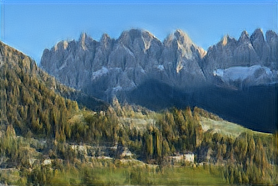}
	\end{subfigure}
	\begin{subfigure}[t]{0.22\textwidth}
		\centering
		\includegraphics[width=1\linewidth,]{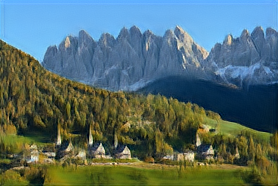}
	\end{subfigure}
	\end{subfigure}
	\begin{subfigure}[t]{0.15\textwidth}
		\centering
		\caption{Sample from id \#2}
		\includegraphics[width=1\linewidth,]{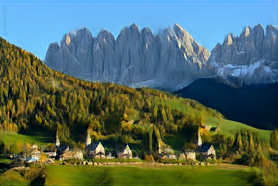}
	\end{subfigure}
	\par\smallskip

	\begin{subfigure}[t]{1\textwidth}
		\centering
		\caption{Spatial sampling}
	\end{subfigure}
	\begin{subfigure}[t]{0.215\textwidth}
		\centering
		\includegraphics[width=1\linewidth,]{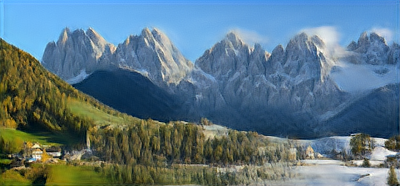}
	\end{subfigure}\hspace{1.2mm}
	\begin{subfigure}[t]{0.15\textwidth}
		\centering
		\includegraphics[width=1\linewidth,]{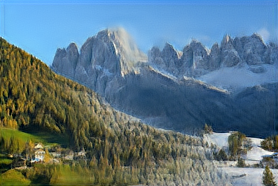}
	\end{subfigure}\hspace{1.2mm}
	\begin{subfigure}[t]{0.215\textwidth}
		\centering
		\includegraphics[width=1\linewidth,]{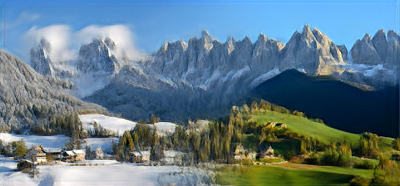}
	\end{subfigure}\hspace{1.2mm}
	\begin{subfigure}[t]{0.215\textwidth}
		\centering
		\includegraphics[width=1\linewidth,]{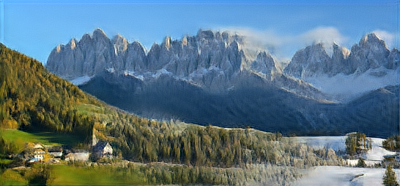}
	\end{subfigure}\hspace{1.2mm}
	\begin{subfigure}[t]{0.15\textwidth}
		\centering
		\includegraphics[width=1\linewidth,]{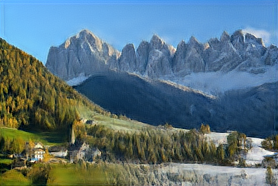}
	\end{subfigure}
	
	\begin{subfigure}[t]{0.215\textwidth}
		\centering
		\frame{\includegraphics[width=1\linewidth,,trim={0 5cm 0 0},clip]{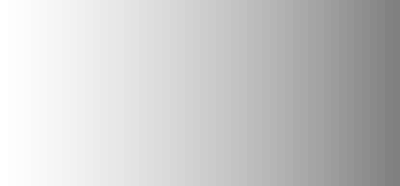}}
	\end{subfigure}\hspace{1.2mm}
	\begin{subfigure}[t]{0.15\textwidth}
		\centering
		\frame{\includegraphics[width=1\linewidth,,trim={0 5cm 0 0},clip]{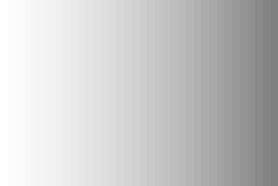}}
	\end{subfigure}\hspace{1.2mm}
	\begin{subfigure}[t]{0.215\textwidth}
		\centering
		\frame{\includegraphics[width=1\linewidth,,trim={0 5cm 0 0},clip]{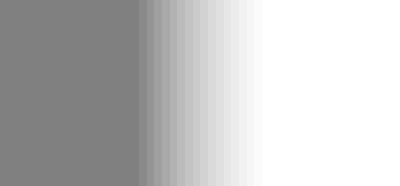}}
	\end{subfigure}\hspace{1.2mm}
	\begin{subfigure}[t]{0.215\textwidth}
		\centering
		\frame{\includegraphics[width=1\linewidth,,trim={0 5cm 0 0},clip]{figures/blending/y400-lin.png}}
	\end{subfigure}\hspace{1.2mm}
	\begin{subfigure}[t]{0.15\textwidth}
		\centering
		\frame{\includegraphics[width=1\linewidth,,trim={0 5cm 0 0},clip]{figures/blending/y278-lin.png}}
	\end{subfigure}

\caption{\textbf{BlendGAN. } Our framework enables to train a single-image GAN model on several single images (two in this example), such that generation is conditioned on the image identity. At test time, our model can realistically mix between the images or between random samples generated from their statistics. Mixture weights can vary \eg throughout space (last row), over time (top rows) or scales (sec.~\ref{sec:exp}).
\vspace{-3mm}
}
\label{fig:figure1}
\end{figure*}{}

%% file: sections/intro.tex
\section{Introduction}

Training an unconditional generative model on a single natural image has drawn significant attention in recent years. This idea was first introduced by SinGAN~\cite{shaham2019singan}, and was quickly followed by improved models such as ConSinGAN~\cite{hinz2021improved}, Patch VAE-GAN~\cite{gur2020hierarchical} and one-shot GAN~\cite{sushko2021one}. After training on a single image, these models can generate diverse samples that resemble that image, and can also be used to edit, manipulate, and even super-resolve the image. The idea of training a GAN on a single signal has seen many uses and extensions, including in domains like medical imaging~\cite{zhang2020learning}, video~\cite{gur2020hierarchical}, audio~\cite{greshler2021catch}, 3D graphics~\cite{hertz2020deep}, and computer game design~\cite{awiszus2020toad}.
However, despite its merits, training a GAN on a single example has a notable limitation; it cannot be used for image manipulation tasks involving multiple images. 

In this paper, we introduce BlendGAN, a method for training an unconditional GAN on a small number of natural images, such that generation is conditioned on the image identity. Specifically, in addition to random noise, BlendGAN accepts an image-identity map that has the same dimensions as the input images. At inference time, this map can be used to direct the model to draw samples corresponding to one particular training image, thus enabling all tasks that single-image-GANs intend to solve on each image separately, with only one single model. But, more importantly, this image-id map enables a wide range of new applications, which are achieved by injecting spatially varying, non-categorical maps. Example applications include morphing, stitching, spatial blending, and style mixing between either the training images or random samples corresponding to their internal statistics. A few such examples are illustrated in Fig.~\ref{fig:figure1}.

Similarly to externally trained GANs, our conditioning mechanism makes use of adaptive normalization, which affects only the denormalization parameters of the normalization units within the model. However, in sharp contrast to external models, our conditioning is not on classes or attributes, which are features shared across many image examples in a large training set. Our conditioning is rather on the identity of each individual training example. Thus, in analogy with single-image-GAN models, which treat a single image as a dataset of patches, here our dataset is the pool of all patches from all images, and subsets of patches belonging to different images are regarded as examples from different ``classes''. It should be noted that our patch-GAN framework operates simultaneously on all overlapping patches of each training image by using a convolutional architecture. Therefore, during training our model is exposed only to spatially constant categorical identity maps. Nevertheless, at inference time, it manages to accommodate spatially varying non-categorical identity structures, which is what enables the many applications of BlendGAN.

In addition to the conditioning mechanism, we also present a modified training procedure, which enables faster training, as well as handling of higher resolution images, without sacrificing image quality. Moreover, we show that increasing the training set to a few images requires only a moderate increase in capacity.

%% file: sections/related.tex
\section{Related work}

\input{figures/vs_naive/vs_naive}

\textbf{Single image GANs}
Training an unconditional generative model on a single image was initially used for texture generation~\cite{bergmann2017learning,jetchev2016texture,li2016precomputed,zhou2018non}. The idea was first extended to general natural images by SinGAN~\cite{shaham2019singan} which utilizes a pyramid of GANs, each of which learns the internal patch distribution of the image at a different scale. This was followed by improved models such as ConSinGAN~\cite{hinz2021improved} which suggests a new training process, Patch VAE-GAN~\cite{gur2020hierarchical} which replaces the GANs of the coarser pyramid scales with patch-VAE, and one-shot GAN~\cite{sushko2021one} which uses two branch discriminator to better control the image content as well as its global layout. Training a GAN model on a single datum was also adapted to several additional domains including medical imaging~\cite{zhang2020learning}, video~\cite{gur2020hierarchical}, audio~\cite{greshler2021catch}, 3D graphics~\cite{hertz2020deep}, and computer game design~\cite{awiszus2020toad}. 

\noindent\textbf{GAN training with limited data} 
Recent works addressed the task of training a GAN model with limited data. This was done either by adapting the weights of a pre-trained model according to several dozens of training examples~\cite{liu2020towards}, or by training a model from scratch~\cite{karras2020training,zhao2020differentiable,cui2021genco,kong2022few}. However, when training such models on an extremely small set of images (\eg two or three images) they usually completely fail to generate realistic results (see Fig.~\ref{fig:vs_naive}). Our model is conditioned on the training image identities and is therefore able to generate plausible samples from each image separately, and to realistically  combine image samples.

\noindent\textbf{Image compositing and melding}
Soimakov \etal~\cite{simakov2008summarizing} presented a method for seamlessly fusing several images into a montage. Their algorithm solves an optimization problem involving a novel bi-directional similarity measure between the images. Pritch \etal~\cite{pritch2009shift} introduced the shift-map algorithm, which is also applicable to this task. This method copies non-rectangular chunks from the input images and seamlessly stitches them. Later, Darabi \etal~\cite{darabi2012image} used patch-based methods to synthesize a smooth transition region between two source images, so that inconsistent color, texture, and structural properties changed gradually from one image to the other. More recently, Yu \etal~\cite{yu2019texture} employed a GAN to create a realistic and smooth interpolation between an arbitrary number of texture samples. Their method, however, is limited to textures. This is while our goal here, is to handle arbitrary natural images.

\noindent\textbf{Image morphing}
Image morphing is the process of interpolating between two images, while maintaining photo-realism. Early morphing techniques involved geometric warping combined with cross-dissolving operations. Beier and Neely \cite{beier1992feature} utilized user-defined line segments to create a dense correspondence map. Other works avoided manual intervention; Liao \etal \cite{liao2014automating} extracted correspondences automatically by optimizing a term similar to structural image similarity. In \cite{shechtman2010regenerative}, the intermediate images were generated by minimizing the bidirectional similarity between neighboring frames. Zhu \etal~\cite{zhu2016generative} projected the images onto the latent space of a pre-trained GAN, and performed linear interpolation in latent space. Simon and Aberdam \cite{Simon_2020_CVPR} later extended this approach by traversing the latent space non-linearly. These latter approaches, however, are constrained to the images that the pre-trained GAN can generate. Our approach involves training only on the particular images we wish to manipulate, and is therefore less constrained.

\noindent\textbf{Adaptive normalization}
Several generative models use a conditional scheme that affects the output by modifying normalization units. In BigGAN~\cite{brock2018large}, class information influences conditional batch normalization (CBN), a class-conditional variant of batch normalization. StyleGAN~\cite{karras2019style} uses an adaptive instance normalization (AdaIN) module, which aligns the mean and variance of the content features with those of the style features. Unlike the CBN, AdaIN does not learn the affine parameters, but rather calculates them based on the style input. The spatially-adaptive normalization (SPADE) model~\cite{park2019semantic}, uses a conditional normalization method for translating label maps to photo-realistic images. Unlike previous methods, SPADE's normalization parameters are not vectors, but rather tensors with spatial dimensions. Jakoel \etal~\cite{jakoel2022gans} use this approach for spatially controlling generation in pre-trained GANs. 

\begin{figure*}[t]
\centering
\includegraphics[width=0.825\textwidth]{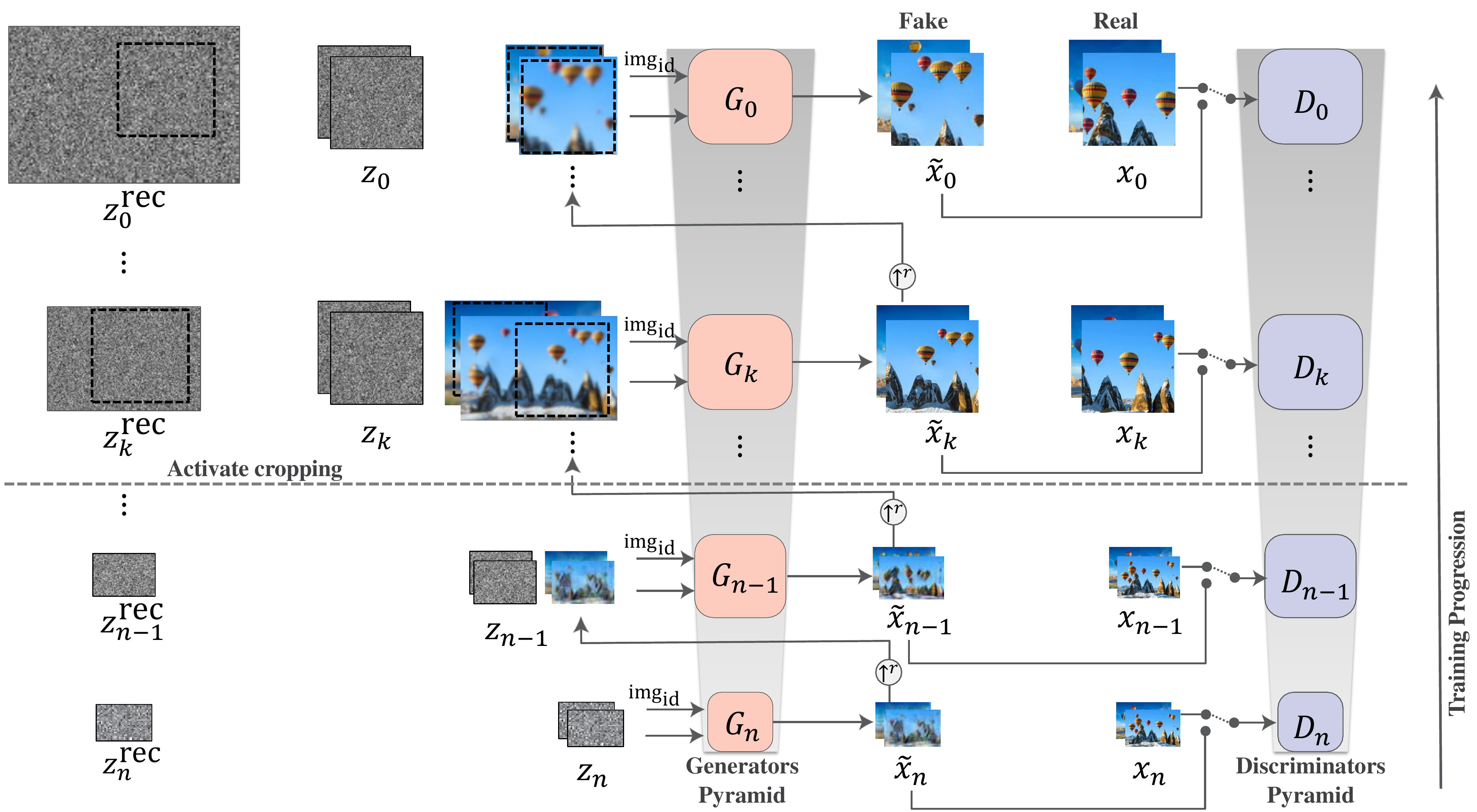}
\caption{\textbf{Architecture.} Our model consists of a pyramid of GANs, each corresponding to a different image scale, similarly to~\cite{shaham2019singan,hinz2021improved,sushko2021one}. To accommodate multiple training images, all generators consume an image identity map that indicates the training image from which statistics to generate. This is fed into SPADE normalization units that modulate the generators' feature maps.}
\label{fig:scheme_multi}
\end{figure*}

%% file: figures/vs_naive/vs_naive.tex
\begin{figure*}[ht]
    \captionsetup[subfigure]{labelformat=empty,justification=centering,aboveskip=1pt,belowskip=1pt}
    \centering
    \begin{tabular}[c]{c c c c}
    \vspace{-1mm}
        \begin{subfigure}[b]{0.122\textwidth}
            \caption*{Training images}
        \end{subfigure} 
        
        &
    
        \begin{subfigure}[b]{0.122\textwidth}
            \caption*{SinGAN}
        \end{subfigure}
        
        \begin{subfigure}[b]{0.122\textwidth}
            \caption*{ConSinGAN}
        \end{subfigure}
        
        &
        
        \begin{subfigure}[b]{0.122\textwidth}
            \caption*{FastGAN}
        \end{subfigure}

        \begin{subfigure}[b]{0.122\textwidth}
            \caption*{MixDL}
        \end{subfigure} 
        
        &

        \begin{subfigure}[b]{0.28975\textwidth}
            \caption*{Ours}
        \end{subfigure}

        \\

        \vspace{-.75mm}
        
        \begin{subfigure}[b]{0.122\textwidth}
            \includegraphics[width=1\linewidth]{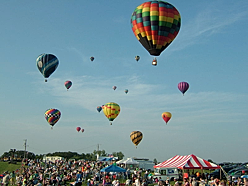}
        \end{subfigure} 
        
        &
    
        \begin{subfigure}[b]{0.122\textwidth}
            \includegraphics[width=1\linewidth]{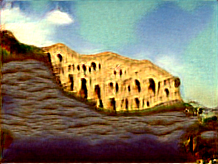}
        \end{subfigure}
        
        \begin{subfigure}[b]{0.122\textwidth}
            \includegraphics[width=1\linewidth]{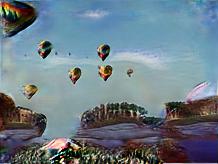}
        \end{subfigure}
        
        &
        
        \begin{subfigure}[b]{0.122\textwidth}
            \includegraphics[width=1\linewidth]{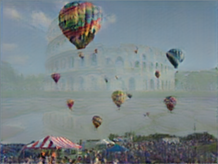}
        \end{subfigure}

        \begin{subfigure}[b]{0.122\textwidth}
            \includegraphics[width=1\linewidth]{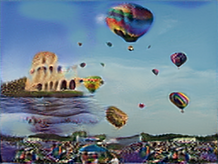}
        \end{subfigure} 
        
        &

        \begin{subfigure}[b]{0.122\textwidth}
            \includegraphics[width=1\linewidth]{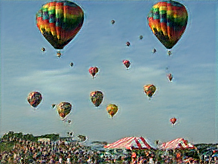}
        \end{subfigure}
        
        \begin{subfigure}[b]{0.16775\textwidth}
            \includegraphics[width=1\linewidth]{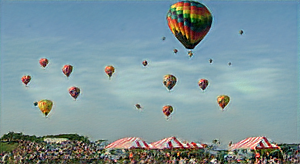}
        \end{subfigure}
        
        
        \\
        
        \begin{subfigure}[b]{0.122\textwidth}
            \includegraphics[width=1\linewidth]{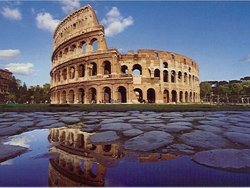}
        \end{subfigure} 
        
        &
    
        \begin{subfigure}[b]{0.122\textwidth}
            \includegraphics[width=1\linewidth]{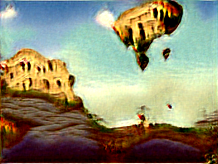}
        \end{subfigure}
        
        \begin{subfigure}[b]{0.122\textwidth}
            \includegraphics[width=1\linewidth]{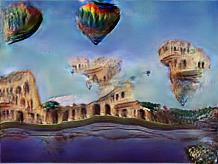}
        \end{subfigure}

        &
        
        \begin{subfigure}[b]{0.122\textwidth}
            \includegraphics[width=1\linewidth]{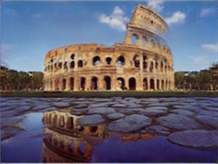}
        \end{subfigure}
        
        \begin{subfigure}[b]{0.122\textwidth}
            \includegraphics[width=1\linewidth]{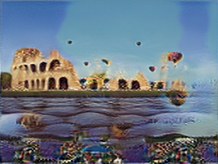}
        \end{subfigure} 
        
        &
    
        \begin{subfigure}[b]{0.122\textwidth}
            \includegraphics[width=1\linewidth]{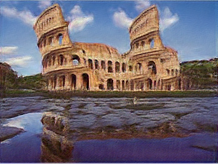}
        \end{subfigure}
        
        \begin{subfigure}[b]{0.16775\textwidth}
            \includegraphics[width=1\linewidth]{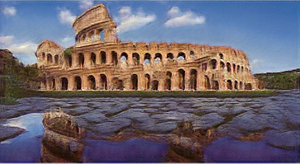}
        \end{subfigure}
        
    
        \\
        &
        \small{Unconditional one shot}
        &
        \small{Unconditional few shot}
        &
        \small{Conditional few shot}
        
    \end{tabular}
    \caption{{\textbf{Two training images.} We trained one-shot GAN models (SinGAN~\cite{shaham2019singan}, ConSinGAN~\cite{hinz2021improved}) 
    on two training images. As can be seen, these models are not designed to deal with more than a single training image. When training few-shot generative models (FastGAN~\cite{liu2020towards}, MixDL~\cite{kong2022few}) on only two images this results in unrealistic combinations of the two. Our model is conditioned on the training image identities and thus at inference is able to generate random samples from each of them separately.
    }}
\label{fig:vs_naive}
\end{figure*}

%% file: sections/method.tex
\section{Method}

\begin{figure*}[t]
\centering
\includegraphics[width=0.85\textwidth]{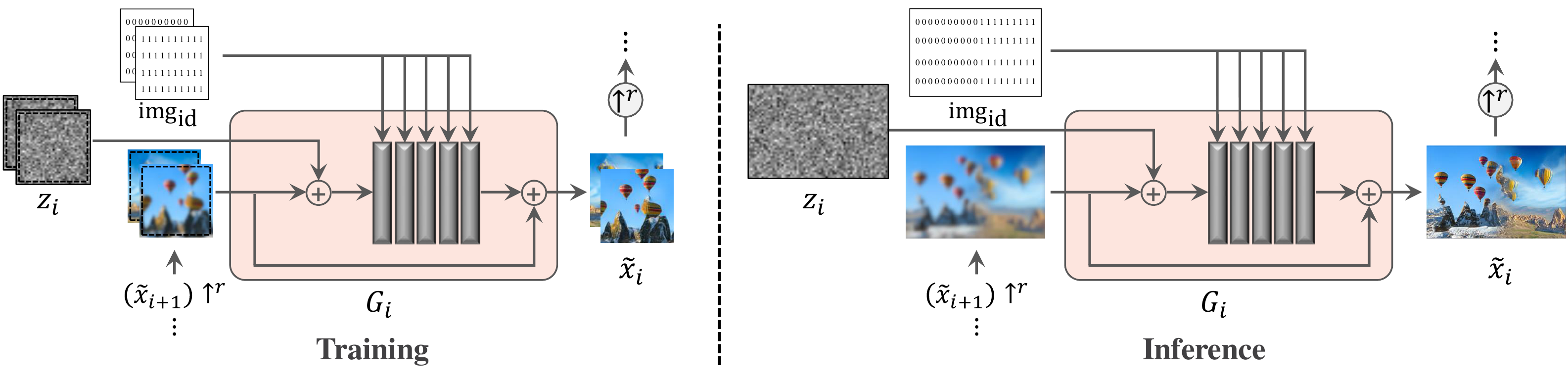}
\caption{\textbf{Training vs. inference.}  
During training, our model is exposed only to categorical, spatially constant, identity maps. At inference, the identity maps can be interpolated over space, or time, or changed across scales, to achieve mixed image identities. 
\vspace{-3mm}
}
\label{fig:single_scale}
\end{figure*}

Single-image GAN models such as~\cite{shaham2019singan,hinz2021improved,sushko2021one} aim to generate fake samples with patches distributed similarly to the patches of the training image, at multiple scales. 
Here, we want to do the same, but for several different images simultaneously. Naively attempting to train one of these models on several images leads to poor results, as can be seen in Fig.~\ref{fig:vs_naive}. Therefore, to achieve our goal, we introduce several important modifications, as illustrated in Fig.~\ref{fig:scheme_multi}. For simplicity we exemplify the use of our approach with the architecture of SinGAN~\cite{shaham2019singan}, however, our method can be easily incorporated into any single image GAN model, as we discuss in the Supplementary Material (SM).

\subsection{Architecture}
We utilize a multi-scale architecture as in \cite{shaham2019singan,hinz2021improved,gur2020hierarchical,sushko2021one}. That is, our model consists of a pyramid of generators, $\{G_0,\ldots,G_n\}$ that are trained to capture the patch statistics within a pyramid $\{x_0,\ldots,x_n\}$ corresponding to each training image~$x$. Here,~$x_i$ is a down-sampled version of~$x$ corresponding to scale factor $r^i$, where~$r$ is the scale factor between consecutive pyramid levels. Each generator $G_i$ is trained against a corresponding patch-discriminator $D_i$. The generator attempts to generate a photo-realistic sample $\widetilde{x}_i$. The discriminator is presented either with the real image $x_i$ or with the fake sample $\widetilde{x}_i$ and tries to classify each of the overlapping patches within its input as real or fake.

All generators in our model consume an image identity tensor, so that this information is available at all scales. At the coarsest scale, $G_n$ takes as input a white Gaussian noise map $z_n$ and an image identity tensor $\text{img}_{\text{id}}$ and outputs a sample $\widetilde{x}_n = G_n(z_n, \text{img}_{\text{id}})$. All other generators additionally consume an up-sampled version of the image generated at the previous scale, so that $\widetilde{x}_i = G_i(z_i, (\widetilde{x}_{i+1})\!\!\uparrow^r,\text{img}_{\text{id}})$ for all $i<n$. All the generators and discriminators are convolutional networks with the same (small) receptive field. Therefore, the coarse scales capture more global structures while the finer scales capture smaller details and textures.

The tensor $\text{img}_{\text{id}}$ has the same spatial dimensions as the image in each scale, and its number of channels equals the number of training images. In each spatial location, it contains a one-hot representation of the identity of the training image from the statistics of which we wish to generate~\cite{park2019semantic}. This tensor goes through two $1\times 1$ convolution layers with ReLU activation to yield the modulation parameters $\gamma$ and $\beta$ for each of the generator's convolutional block. Unlike the conventional batch normalization~\cite{ioffe2015batch}, here $\gamma$ and $\beta$ are tensors with spatial dimensions that are multiplied by and added to the normalized activations, element-by-element. 

\input{figures/texture-mixer/texture_mixer}
\input{figures/sampling-with-map/sampling-with-map.tex}

\subsection{Training}
As in \cite{shaham2019singan,gur2020hierarchical,sushko2021one} training is done scale-by-scale, coarse to fine, while keeping all other scales fixed. The training loss consists of an adversarial term and a reconstruction term. The adversarial loss is that of a Wasserstein GAN with gradient penalty \cite{gulrajani2017improved}. The reconstruction loss guarantees that each training example can be generated by inputting a fixed set of noise maps $\{z_{n}^{\text{rec}}, z_{n-1}^{\text{rec}},..,z_{0}^{\text{rec}} \}$, which are drawn once and kept fixed during training. Importantly, these noise maps are shared by all training images. Thus, generation of each training image can be achieved by injecting the same reconstruction noise maps, but with the image's particular id. To enable the generators to comply with this requirement, here we do not take $z_{n-1}^{\text{rec}},..,z_{0}^{\text{rec}}$ to be zero maps as in \cite{shaham2019singan}. Instead, we take the variance of each $z_{i}^{\text{rec}}$ to be proportional to that of the generation noise $z_{i}$, as computed in the original SinGAN framework. One of the key features of our shared reconstruction noise maps, is that they allow to manipulate the training images themselves at inference time, and not only random samples.

To accommodate large training images without increasing memory consumption, we employ a random cropping scheme at the finer scales. As illustrated in Fig.~\ref{fig:scheme_multi}, we do this for the input image $(\widetilde{x}_{i+1})\uparrow^r$, the random noise $z_i$, and the reconstruction noise $z_{i}^{\text{rec}}$.
This is done by removing the padding from the generator's convolution layers and cropping a region that is larger by half a receptive field from each side. To retain the positional encoding induced by the zero-padding at the borders of the whole image, the full noise tensors $z_{i}^{\text{rec}}$ and $z_i$ are first padded with zeros, so that a crop may occasionally contain part of the zero-padded borders. For small training images, we set the cropping window to $128\times 128$, and for larger ones to $256\times 256$. The cropping allows us to train on significantly larger images than what is usually possible with single-image-GAN models, for a given memory consumption budget. Please see the additional evaluation of our cropping mechanism in the SM, along with examples for high resolution image synthesis.

%% file: figures/texture-mixer/texture_mixer.tex
\begin{figure*}[t]
	\centering
	\captionsetup[subfigure]{labelformat=empty,justification=centering,aboveskip=1pt,belowskip=1pt}
	\begin{subfigure}[t]{0.09\textwidth}
	    \vspace{-.3mm}
		\centering
		\caption{Image \# 1}
		\includegraphics[width=1\linewidth]{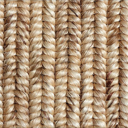}
	\end{subfigure}\hspace{0.5mm}
	\begin{subfigure}[t]{0.09\textwidth}
	    \vspace{-.3mm}
		\centering
		\caption{}
		\includegraphics[width=1\linewidth]{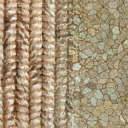}
	\end{subfigure}
	\begin{subfigure}[t]{0.27\textwidth}
	    \vspace{-.3mm}
		\centering
		\caption{Texture Mixer}
		\includegraphics[width=1\linewidth]{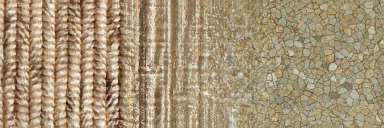}
	\end{subfigure}\hspace{0.5mm}
	\begin{subfigure}[t]{0.09\textwidth}
	    \vspace{-.3mm}
		\centering
		\caption{}
		\includegraphics[width=1\linewidth]{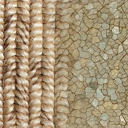}
	\end{subfigure}
	\begin{subfigure}[t]{0.27\textwidth}
	    \vspace{-.3mm}
		\centering
		\caption{Ours}
		\includegraphics[width=1\linewidth]{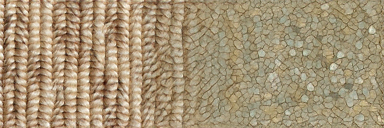}
	\end{subfigure}\hspace{0.5mm}
	\begin{subfigure}[t]{0.09\textwidth}
	    \vspace{-.3mm}
		\centering
		\caption{Image \# 2}
		\includegraphics[width=1\linewidth]{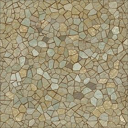}
	\end{subfigure}
	\par\smallskip
	\begin{subfigure}[t]{0.09\textwidth}
	    \vspace{-.3mm}
		\centering
		\includegraphics[width=1\linewidth]{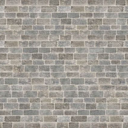}
	\end{subfigure}\hspace{0.5mm}
	\begin{subfigure}[t]{0.09\textwidth}
	    \vspace{-.3mm}
		\centering
		\includegraphics[width=1\linewidth]{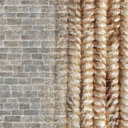}
	\end{subfigure}
	\begin{subfigure}[t]{0.27\textwidth}
	    \vspace{-.3mm}
		\centering
		\includegraphics[width=1\linewidth]{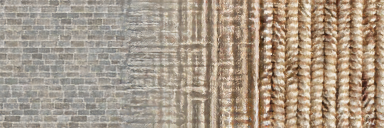}
	\end{subfigure}\hspace{0.5mm}
	\begin{subfigure}[t]{0.09\textwidth}
	    \vspace{-.3mm}
		\centering
		\includegraphics[width=1\linewidth]{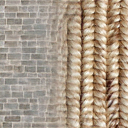}
	\end{subfigure}
	\begin{subfigure}[t]{0.27\textwidth}
	    \vspace{-.3mm}
		\centering
		\includegraphics[width=1\linewidth]{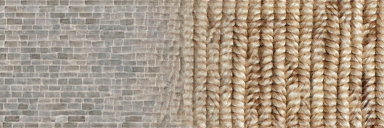}
	\end{subfigure}\hspace{0.5mm}
	\begin{subfigure}[t]{0.09\textwidth}
	    \vspace{-.3mm}
		\centering
		\includegraphics[width=1\linewidth]{figures/texture-mixer/images/paving2.png}
	\end{subfigure}
	\par\smallskip
	\begin{subfigure}[t]{0.09\textwidth}
	    \vspace{-.3mm}
		\centering
		\includegraphics[width=1\linewidth]{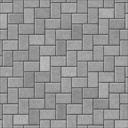}
	\end{subfigure}\hspace{0.5mm}
	\begin{subfigure}[t]{0.09\textwidth}
	    \vspace{-.3mm}
		\centering
		\includegraphics[width=1\linewidth]{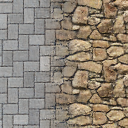}
	\end{subfigure}
	\begin{subfigure}[t]{0.27\textwidth}
	    \vspace{-.3mm}
		\centering
		\includegraphics[width=1\linewidth]{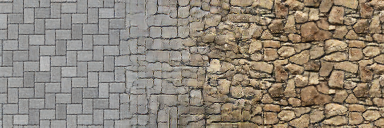}
	\end{subfigure}\hspace{0.5mm}
	\begin{subfigure}[t]{0.09\textwidth}
	    \vspace{-.3mm}
		\centering
		\includegraphics[width=1\linewidth]{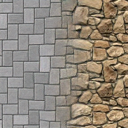}
	\end{subfigure}
	\begin{subfigure}[t]{0.27\textwidth}
	    \vspace{-.3mm}
		\centering
		\includegraphics[width=1\linewidth]{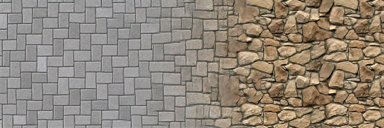}
	\end{subfigure}\hspace{0.5mm}
	\begin{subfigure}[t]{0.09\textwidth}
	    \vspace{-.3mm}
		\centering
		\includegraphics[width=1\linewidth]{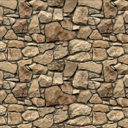}
	\end{subfigure}
	\par\smallskip
	\begin{subfigure}[t]{0.09\textwidth}
	    \vspace{-.3mm}
		\centering
		\includegraphics[width=1\linewidth]{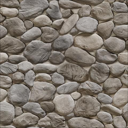}
	\end{subfigure}\hspace{0.5mm}
	\begin{subfigure}[t]{0.09\textwidth}
	    \vspace{-.3mm}
		\centering
		\includegraphics[width=1\linewidth]{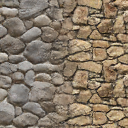}
	\end{subfigure}
	\begin{subfigure}[t]{0.27\textwidth}
	    \vspace{-.3mm}
		\centering
		\includegraphics[width=1\linewidth]{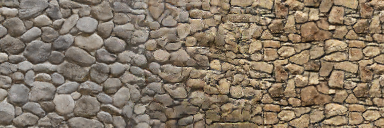}
	\end{subfigure}\hspace{0.5mm}
	\begin{subfigure}[t]{0.09\textwidth}
	    \vspace{-.3mm}
		\centering
		\includegraphics[width=1\linewidth]{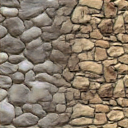}
	\end{subfigure}
	\begin{subfigure}[t]{0.27\textwidth}
	    \vspace{-.3mm}
		\centering
		\includegraphics[width=1\linewidth]{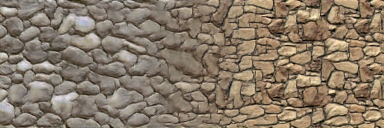}
	\end{subfigure}\hspace{0.5mm}
	\begin{subfigure}[t]{0.09\textwidth}
	    \vspace{-.3mm}
		\centering
		\includegraphics[width=1\linewidth]{figures/texture-mixer/images/paving4.png}
	\end{subfigure}

	\caption{\label{fig:textures_blending}{\textbf{Image melding.} Comparison with the state-of-the-art Texture-Mixer. 
	Our model generates a plausibly looking, smoothly varying transition region between the two sources, while Texture-Mixer synthesizes intermediate textures that do not resemble any of the sources.
}
\vspace{-5mm}
}
\end{figure*}{}

%% file: figures/sampling-with-map/sampling-with-map.tex

    
\begin{figure*}[ht]
\centering
\captionsetup[subfigure]{labelformat=empty,justification=centering,aboveskip=1pt,belowskip=1pt}
    \begin{tabular}[c]{c c}

    \centering
    
    \begin{subfigure}[t]{0.18\textwidth}
    \centering
    \caption*{Training images}
    \end{subfigure}
    
    &
    
    \begin{subfigure}[t]{0.777\textwidth}
    \centering
    \caption*{Random samples}
    \end{subfigure}

    \\
        
    \begin{subfigure}[t]{0.18\textwidth}
    \centering
    \includegraphics[width=1\linewidth]{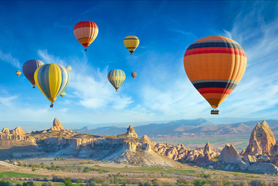}
    \end{subfigure}

    &

    \begin{subfigure}[t]{0.259\textwidth}
    \centering
    \includegraphics[width=1\linewidth]{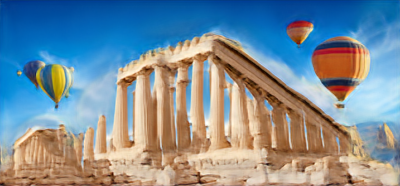}
    \end{subfigure}

    \begin{subfigure}[t]{0.259\textwidth}
    \centering
    \includegraphics[width=1\linewidth]{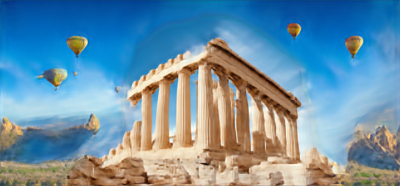}
    \end{subfigure}
    
    \begin{subfigure}[t]{0.259\textwidth}
    \centering
    \includegraphics[width=1\linewidth]{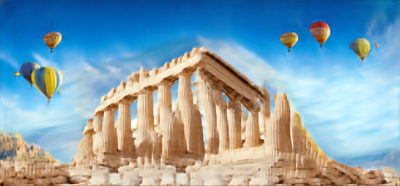}
    \end{subfigure}

    \\

        
    \begin{subfigure}[t]{0.18\textwidth}
    \centering
    \includegraphics[width=1\linewidth]{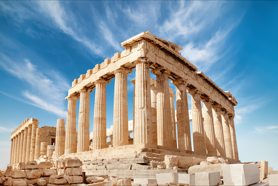}
    \end{subfigure}

    &

    \begin{subfigure}[t]{0.259\textwidth}
    \centering
    \includegraphics[width=1\linewidth]{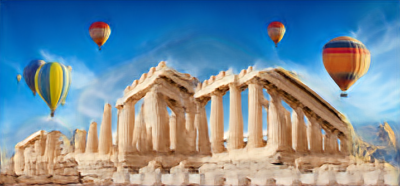}
    \end{subfigure}

    \begin{subfigure}[t]{0.259\textwidth}
    \centering
    \includegraphics[width=1\linewidth]{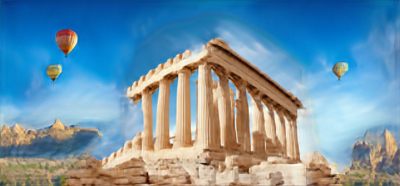}
    \end{subfigure}

    \begin{subfigure}[t]{0.259\textwidth}
    \centering
    \includegraphics[width=1\linewidth]{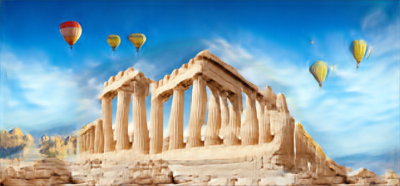}
    \end{subfigure}

    \\

    \begin{subfigure}[t]{0.18\textwidth}
    \centering
    \vspace{-12mm}
    \caption*{Image-identity}
    \end{subfigure}
    
    &
    
    \begin{subfigure}[t]{0.259\textwidth}
    \centering
    \frame{\includegraphics[width=1\linewidth]{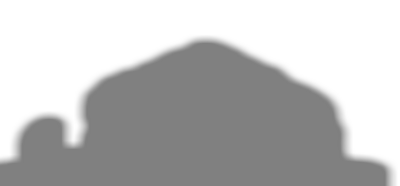}}
    \end{subfigure}

    \begin{subfigure}[t]{0.259\textwidth}
    \centering
    \frame{\includegraphics[width=1\linewidth]{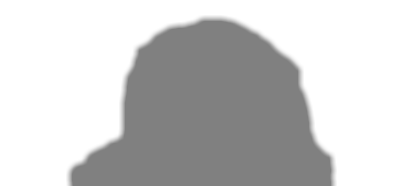}}
    \put(-85, 15){\footnotesize{Reconstruction}}
    \end{subfigure}

    \begin{subfigure}[t]{0.259\textwidth}
    \centering
    \frame{\includegraphics[width=1\linewidth]{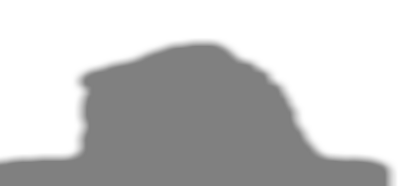}}
    \end{subfigure}

    \end{tabular}
\caption{\textbf{2D Spatial sampling.} At inference, our model accommodates spatially varying non-categorical identity structures, allowing the user to choose what image identity to sample from at each 
pixel. 
This can be done with random samples (right and left columns) 
or with the reconstruction noise such that part of the image remain faithful with the training image (\eg the Pantheon region in the central column).
\vspace{-5mm}
}
\label{fig:sampling_with_map}
\end{figure*}{}

%% file: sections/experiments.tex
\section{Experiments}\label{sec:exp}

\input{figures/morphing/moprhing}
\input{figures/structure-texture/st-tex}
\input{figures/multi/balloons-birds-colusseum-stone/multi.tex}
We illustrate the utility of BlendGAN in several tasks, which we address by varying the image-id at inference time across space (melding), time (morphing), and scale (structure-texture fusion). We also provide an evaluation of the generation performance. Please see additional results in the SM.

\subsection{Image Melding}
Image melding, coined in \cite{darabi2012image}, refers to the task of synthesizing a transition region between two source images, such that colors, textures, and structures change gradually from one source to the other. Although our model is trained using only spatially constant categorical identity maps, we can perform such manipulations at inference time by injecting spatially varying non-categorical identity maps. To constrain the image to match the source images on both sides, we use the reconstruction noise maps at those two locations. To synthesize an output image that is larger than the two inputs, we use random noise in between the two reconstruction noise maps.

Figure \ref{fig:textures_blending} shows a comparison with Texture Mixer \cite{yu2019texture} for the task of interpolating two texture images. In this case, we set  the transition region to be $1/3$ of the output image. As can be seen, our method generates a plausibly looking gradual transition between the images, while Texture Mixture synthesizes intermediate textures that do not resemble either of the sources. Please refer to the SM for melding between three images with our model.

As opposed to Texture Mixer, our method is not limited to texture images. Furthermore, it can also be used to blend \emph{random samples} rather than the images themselves. Figure~\ref{fig:figure1} illustrates our model's ability to spatially blend natural images. Particularly, here we show various spatial transition patterns between random samples generated by our model. Figure~\ref{fig:sampling_with_map} illustrates sharper transitions between regions in the shapes of user-prescribed masks. As can be seen this allows to generate samples that realistically combine the content from both images. 

\begin{figure}[t]
	\centering
	\captionsetup[subfigure]{labelformat=empty,justification=centering,aboveskip=1pt,belowskip=1pt}
	\begin{subfigure}[t]{0.5\textwidth}
	    \vspace{-.3mm}
		\centering
		\includegraphics[width=0.675\linewidth]{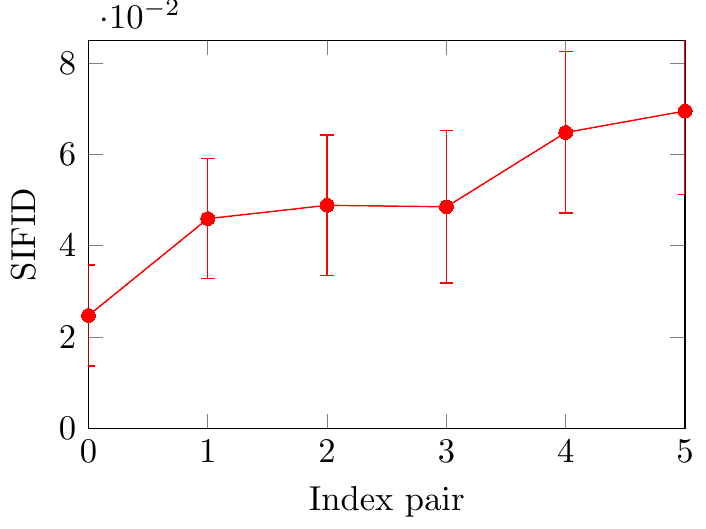}
	\end{subfigure}
	\begin{subfigure}[t]{0.5\textwidth}
	    \vspace{-.3mm}
		\centering
		\includegraphics[width=0.9\linewidth]{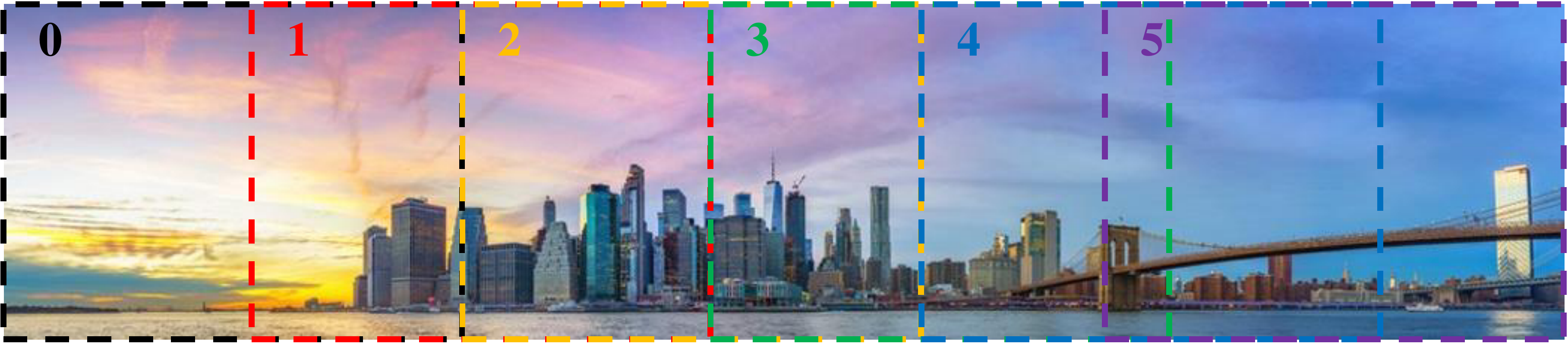}
	\end{subfigure}
	\caption{\textbf{Quality as a function of the training images diversity.} We train a BlendGAN model on two images: crop $0$ and crop $i$, with $i=1,\ldots,5$. For each trained model, SIFID scores are calculated on samples corresponding to id $0$. As expected, the more similar the training images, the better the quality.}
\label{fig:pano}
\end{figure}

\begin{figure}[t]
\centering
\includegraphics[width=.775\linewidth]{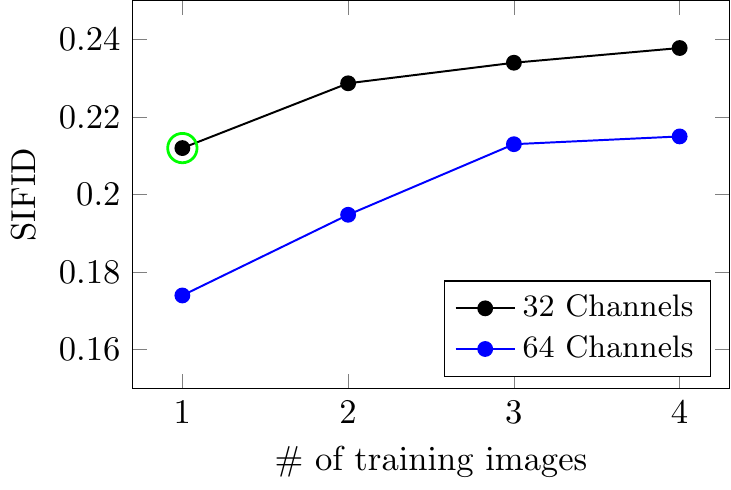}
\caption{\textbf{Multiple training images.} Two variants of our model trained on an increasing number of training images. Our BlendGAN can accommodate four training images while maintaining similar quality to the original SinGAN (green dot circle).
}
\label{fig:capacity}
\end{figure}

\subsection{Image Morphing}
The goal in image morphing is to interpolate between two images, while maintaining photo-realism. Here, we accomplish this task by interpolating the id's of the two images continuously over time. When injecting random noise, we obtain morphing between random samples, and when using the reconstruction noise we morph between the images themselves.

Figures \ref{fig:figure1} and \ref{fig:morphing} depict morphing results generated by our method. Our models are trained for each pair of natural images, using spatially constant categorical identity maps. Then at inference time, we generate interpolated images with image-id values of $\{0.2, 0.4, 0.6, 0.8\}$. As can be seen, or model successfully generalizes to non-categorical id values, managing to generate, for example, plausible crossbreeds between strawberries and raspberries. Please see additional morphing results including three images in the SM.

\input{figures/clip/clip}

\subsection{Structure-Texture Fusion}
BlendGAN can be used to fuse textures and structures taken from two images, as illustrated in Fig.~\ref{fig:structure-texture}. At inference, we inject id-maps corresponding to the structure image at the coarser scales of the model and id-maps corresponding to the texture image at finer scales. This leads to generation of images with structures resembling those of the first image and textures resembling those of the second image. The scale at which we transition from one id to the other determines the sizes of image elements taken from each of them.

\subsection{Image Generation Analysis}
Figures \ref{fig:vs_naive},\ref{fig:multi} shows random samples generated by BlendGAN trained on multiple images (two images in Fig.~\ref{fig:vs_naive} and four images in Fig.~\ref{fig:multi}). For each image id we present several random samples having different dimensions and aspect ratios. As can be seen, our model generates new realistic structures, while maintaining semantic similarity to the training images. Note, that our model is only trained once over all the images, and can be used at inference time on each of them separately. This is in contrast to single-image GAN models, where a distinct model should be trained for each individual image.  
In the SM we illustrate how such a model can be used to perform all the editing and manipulation tasks that the vanilla SinGAN model (trained on each image individually) can perform.

\myparagraph{Quality as function of diversity.}
When training a limited capacity model on multiple images, it is natural to expect that the quality of the samples depend on the diversity of textures and structures between the training images. Figure~\ref{fig:pano} illustrates this dependence. Here, we take a panoramic image and crop six partially overlapping sub-images. We train our model on pairs of sub-images consisting of the leftmost crop, indexed as $0$, and one additional crop. For each trained model we measure the average SIFID between crop $0$ and fifty random samples generated with id $0$. As can be seen, the generation process is of higher quality when the two training images exhibit stronger similarities.

\myparagraph{Quality as function of the number of training images.}
A complementary question, is how the number of training images affects the generation quality. This is explored in Fig.~\ref{fig:capacity}. Here, we train two variants of our model using a varying number of training examples from SinGAN's repository (Balloons and images $1,2,4,5,10,11,50$). The first model has the same capacity as the original SinGAN with $32$ channels at the coarser scales (black), while the second model has $64$ channels (blue). A different pair of images is used for training each model separately five times. Next, we measure the average SIFID between the \textit{Balloons} image and fifty samples generated by the \textit{Balloons} id, and average over all five models. As can be seen, BlendGAN can accommodate four training images while maintaining similar quality to the original SinGAN (green circle).

\subsection{Closing the Gap Between Internal Distributions}

Training with only on a few single images limits the ability of BlendGAN to capture complex semantic context. That is, at inference, BlendGAN may produce unrealistic artifacts for non-integer identity maps, especially when the training images contain distinct semantic contents (Fig.~\ref{fig:clip}, first row). Note however that this can be mitigated by adding an external semantic blending loss 

\begin{equation}\label{eq:full_ref_measure}
    {\cal L}_\text{semantics}(z^{\text{rec}},{r}^{\text{id}}) = \| \phi (G(z^{\text{rec}}, r^{\text{id}})) - \sum_{i} \alpha_i  \phi(\text{img}_i)\|_1.
\end{equation}
Here, $\{\alpha_i\}$ are randomly drawn constants that sum to 1, $r^{\text{id}}$ is a spatially-constant image-id map with weights $\{\alpha_i\}$ (rather than a one-hot vector), and $\phi$ is the CLIP embedding~\cite{radford2021learning} that provides a representation for the images' semantic contents. That is, during training we enforce the blending result to have an embedding that corresponds to the weighted average of the embedding of the training images, according to the drawn identity map. 
In Fig.~\ref{fig:clip} we show the results of a model trained with and without the semantic blending constraint. Incorporating ${\cal L}_\text{semantics}$ to the training results in sharper generation with less artifacts.

%% file: figures/morphing/moprhing.tex
\begin{figure*}[ht]
	\centering
	\captionsetup[subfigure]{labelformat=empty,justification=centering,aboveskip=1pt,belowskip=1pt}
	\begin{subfigure}[t]{0.15\textwidth}
	    \vspace{-.3mm}
		\centering
		\caption{Image \# 1}
		\includegraphics[width=1\linewidth]{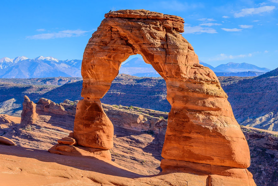}
	\end{subfigure}
	\begin{subfigure}[t]{0.69\textwidth}
	\centering
	\caption{Image morphing}
	\begin{subfigure}[t]{0.22\textwidth}
	    \vspace{-.3mm}
		\centering
		\includegraphics[width=1\linewidth]{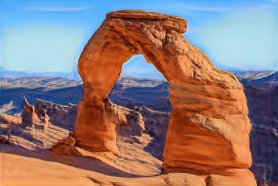}
	\end{subfigure}
	\begin{subfigure}[t]{0.22\textwidth}
	    \vspace{-.3mm}
		\centering
		\includegraphics[width=1\linewidth]{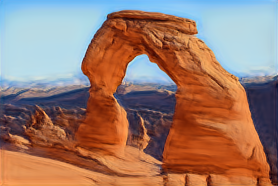}
	\end{subfigure}
	\begin{subfigure}[t]{0.22\textwidth}
	    \vspace{-.3mm}
		\centering
		\includegraphics[width=1\linewidth]{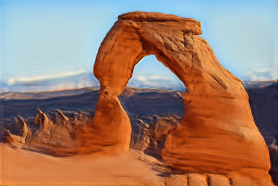}
	\end{subfigure}
	\begin{subfigure}[t]{0.22\textwidth}
	    \vspace{-.3mm}
		\centering
		\includegraphics[width=1\linewidth]{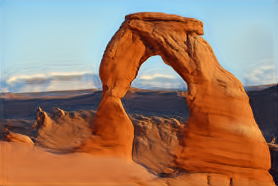}
	\end{subfigure}
	\end{subfigure}
	\begin{subfigure}[t]{0.15\textwidth}
	    \vspace{-.3mm}
		\centering
		\caption{Image \# 2}
		\includegraphics[width=1\linewidth]{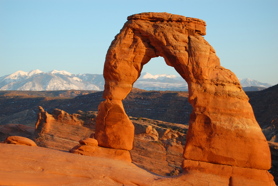}
	\end{subfigure}
	\par\smallskip
	
	\begin{subfigure}[t]{0.15\textwidth}
	    \vspace{-.3mm}
		\centering
		\includegraphics[width=1\linewidth]{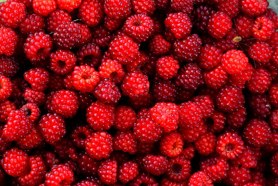}
	\end{subfigure}
	\begin{subfigure}[t]{0.69\textwidth}
	\centering
	\begin{subfigure}[t]{0.22\textwidth}
	    \vspace{-.3mm}
		\centering
		\includegraphics[width=1\linewidth]{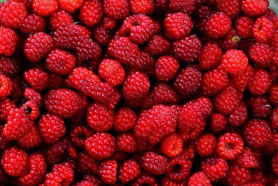}
	\end{subfigure}
	\begin{subfigure}[t]{0.22\textwidth}
	    \vspace{-.3mm}
		\centering
		\includegraphics[width=1\linewidth]{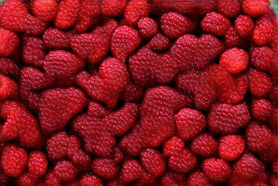}
	\end{subfigure}
	\begin{subfigure}[t]{0.22\textwidth}
	    \vspace{-.3mm}
		\centering
		\includegraphics[width=1\linewidth]{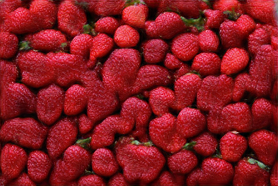}
	\end{subfigure}
	\begin{subfigure}[t]{0.22\textwidth}
	    \vspace{-.3mm}
		\centering
		\includegraphics[width=1\linewidth]{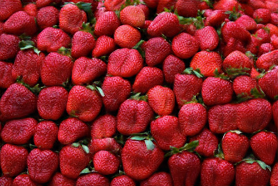}
	\end{subfigure}
	\end{subfigure}
	\begin{subfigure}[t]{0.15\textwidth}
	    \vspace{-.3mm}
		\centering
		\includegraphics[width=1\linewidth]{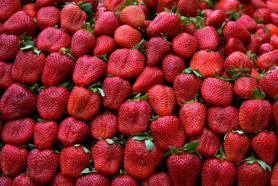}
	\end{subfigure}
	\par\smallskip
	
	\begin{subfigure}[t]{0.15\textwidth}
	    \vspace{-.3mm}
		\centering
		\caption{}
	\end{subfigure}
	\begin{subfigure}[t]{0.69\textwidth}
	\centering
	\caption{}
	\end{subfigure}
	\begin{subfigure}[t]{0.15\textwidth}
	    \vspace{-.3mm}
		\centering
		\caption{}
	\end{subfigure}
    \vspace{-5mm}
    \begin{subfigure}[t]{0.15\textwidth}
	    \vspace{-5mm}
		\centering
		\caption{Sample \# 1}
	\end{subfigure}
	\begin{subfigure}[t]{0.69\textwidth}
	\centering
	\vspace{-5mm}
	\caption{Sample morphing}
	\end{subfigure}
	\begin{subfigure}[t]{0.15\textwidth}
	    \vspace{-5mm}
		\centering
		\caption{Sample \# 2}
	\end{subfigure}
	\par\smallskip

    \begin{subfigure}[t]{0.15\textwidth}
	    \vspace{-1mm}
		\centering
		\includegraphics[width=1\linewidth]{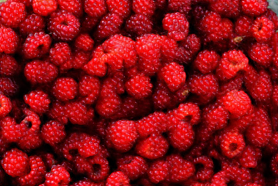}
	\end{subfigure}
	\begin{subfigure}[t]{0.69\textwidth}
	\centering
	\begin{subfigure}[t]{0.22\textwidth}
	    \vspace{-1mm}
		\centering
		\includegraphics[width=1\linewidth]{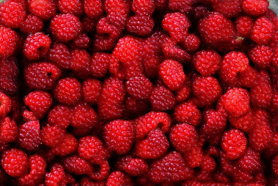}
	\end{subfigure}
	\begin{subfigure}[t]{0.22\textwidth}
	    \vspace{-1mm}
		\centering
		\includegraphics[width=1\linewidth]{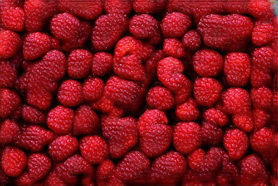}
	\end{subfigure}
	\begin{subfigure}[t]{0.22\textwidth}
	    \vspace{-1mm}
		\centering
		\includegraphics[width=1\linewidth]{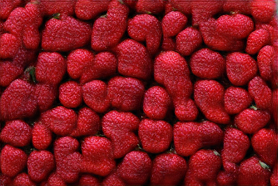}
	\end{subfigure}
	\begin{subfigure}[t]{0.22\textwidth}
	    \vspace{-1mm}
		\centering
		\includegraphics[width=1\linewidth]{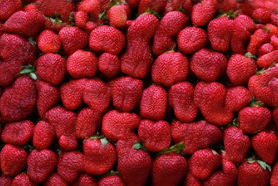}
	\end{subfigure}
	\end{subfigure}
	\begin{subfigure}[t]{0.15\textwidth}
	    \vspace{-1mm}
		\centering
		\includegraphics[width=1\linewidth]{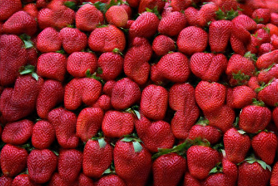}
	\end{subfigure}
	\par\smallskip

   \begin{subfigure}[t]{0.15\textwidth}
	    \vspace{-.3mm}
		\centering
		\includegraphics[width=1\linewidth]{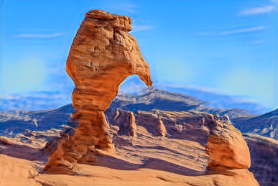}
	\end{subfigure}
	\begin{subfigure}[t]{0.69\textwidth}
	\centering
	\begin{subfigure}[t]{0.22\textwidth}
	    \vspace{-.3mm}
		\centering
		\includegraphics[width=1\linewidth]{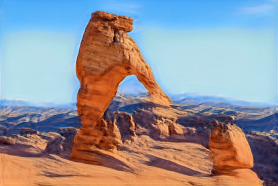}
	\end{subfigure}
	\begin{subfigure}[t]{0.22\textwidth}
	    \vspace{-.3mm}
		\centering
		\includegraphics[width=1\linewidth]{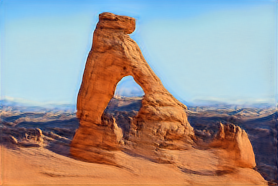}
	\end{subfigure}
	\begin{subfigure}[t]{0.22\textwidth}
	    \vspace{-.3mm}
		\centering
		\includegraphics[width=1\linewidth]{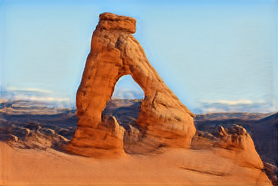}
	\end{subfigure}
	\begin{subfigure}[t]{0.22\textwidth}
	    \vspace{-.3mm}
		\centering
		\includegraphics[width=1\linewidth]{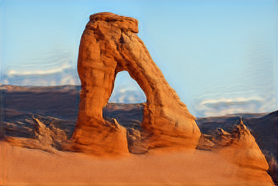}
	\end{subfigure}
	\end{subfigure}
	\begin{subfigure}[t]{0.15\textwidth}
	    \vspace{-.3mm}
		\centering
		\includegraphics[width=1\linewidth]{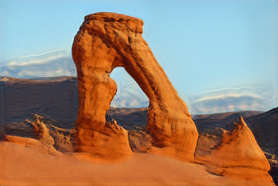}
	\end{subfigure}
	\par\smallskip

	\caption{\label{fig:morphing}{\textbf{Image morphing and sample morphing.} After training on a pair of natural images, image morphing and sample morphing can be achieved by 
	changing the spatially constant image-id maps 
	gradually 
	from the id of one image to the id of the other image over time.}
	}
\end{figure*}{}

%% file: figures/structure-texture/st-tex.tex
\begin{figure*}[ht]
	\centering
	\captionsetup[subfigure]{labelformat=empty,justification=centering,aboveskip=1pt,belowskip=1pt}
	\begin{subfigure}[t]{0.12\textwidth}
		\centering
		\caption{Structure}
    	\includegraphics[width=1\linewidth]{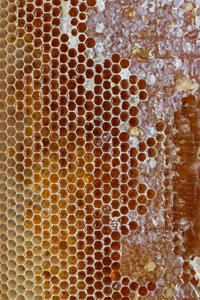}
    \end{subfigure}
	\begin{subfigure}[t]{0.6\textwidth}
	\centering
	\caption{Structure-texture fusion}
	\hspace{-1mm}
    	\begin{subfigure}[t]{0.2\textwidth}
    		\centering
    		\includegraphics[width=1\linewidth]{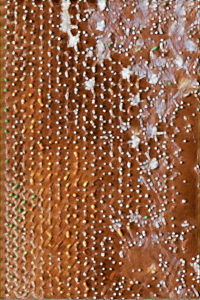}
        \end{subfigure}
        \begin{subfigure}[t]{0.2\textwidth}
    		\centering
    		\includegraphics[width=1\linewidth]{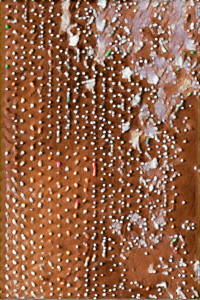}
        \end{subfigure}
        \begin{subfigure}[t]{0.2\textwidth}
    		\centering
    		\includegraphics[width=1\linewidth]{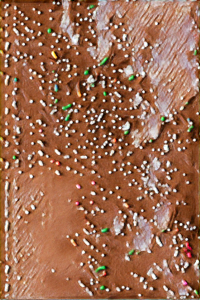}
        \end{subfigure}
        \begin{subfigure}[t]{0.2\textwidth}
    		\centering
    		\includegraphics[width=1\linewidth]{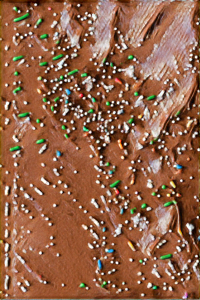}
        \end{subfigure}
    \end{subfigure}
    \begin{subfigure}[t]{0.12\textwidth}
		\centering
		\caption{Texture}
		\hspace{-1mm}
    	\includegraphics[width=1\linewidth]{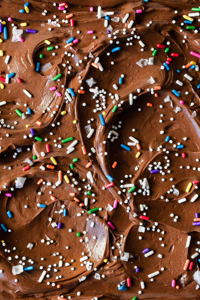}
    \end{subfigure}


	\begin{subfigure}[t]{0.12\textwidth}
		\centering
		\caption{}
    \end{subfigure}
	\begin{subfigure}[t]{0.6\textwidth}
	\centering
	\hspace{-1mm}
    	\begin{subfigure}[t]{0.2\textwidth}
    		\centering
    		\caption{Scale \#8}
        \end{subfigure}
        \begin{subfigure}[t]{0.2\textwidth}
    		\centering
    		\caption{Scale \#6}
        \end{subfigure}
        \begin{subfigure}[t]{0.2\textwidth}
    		\centering
    		\caption{Scale \#4}
        \end{subfigure}
        \begin{subfigure}[t]{0.2\textwidth}
    		\centering
    		\caption{Scale \#2}
        \end{subfigure}
    \end{subfigure}
    \begin{subfigure}[t]{0.12\textwidth}
		\centering
		\caption{}
		\hspace{-1mm}
    \end{subfigure}
    
\caption{\textbf{Structure-Texture Fusion.} BlendGAN can be used to fuse structure and texture from different images. At inference we use id maps corresponding to the structure image at the generator's coarser scales and to the texture image at finer scales. The number below the images indicate the scale in which the transition was done, which determines the sizes of image elements taken from each of the images.}
\label{fig:structure-texture}
\end{figure*}{}

%% file: figures/multi/balloons-birds-colusseum-stone/multi.tex
\begin{figure*}[t]
\centering
\captionsetup[subfigure]{labelformat=empty,justification=centering,aboveskip=1pt,belowskip=1pt}
    \begin{tabular}[c]{c c}

    \centering
    
    \begin{subfigure}[t]{0.12\textwidth}
    \centering
    \caption*{Training images}
    \end{subfigure}
    
    &
    
    \begin{subfigure}[t]{0.72\textwidth}
    \centering
    \caption*{Random samples}
    \end{subfigure}

    \\
        
    \begin{subfigure}[t]{0.12\textwidth}
    \centering
    \includegraphics[width=1\linewidth]{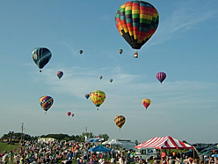}
    \end{subfigure}

    &

    \begin{subfigure}[t]{0.24\textwidth}
    \centering
    \includegraphics[width=1\linewidth]{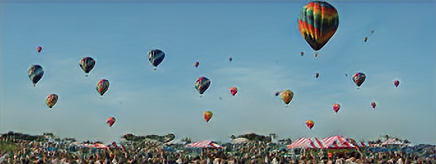}
    \end{subfigure}

    \begin{subfigure}[t]{0.12\textwidth}
    \centering
    \includegraphics[width=1\linewidth]{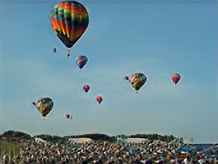}
    \end{subfigure}

    \begin{subfigure}[t]{0.12\textwidth}
    \centering
    \includegraphics[width=1\linewidth]{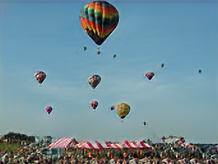}
    \end{subfigure}

    \begin{subfigure}[t]{0.24\textwidth}
    \centering
    \includegraphics[width=1\linewidth]{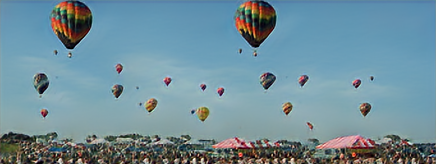}
    \end{subfigure}

    \\
        
    \begin{subfigure}[t]{0.12\textwidth}
    \centering
    \includegraphics[width=1\linewidth]{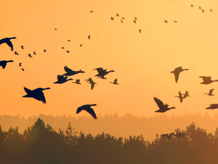}
    \end{subfigure}

    &

    \begin{subfigure}[t]{0.12\textwidth}
    \centering
    \includegraphics[width=1\linewidth]{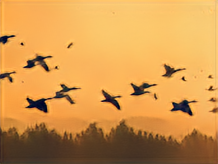}
    \end{subfigure}

    \begin{subfigure}[t]{0.24\textwidth}
    \centering
    \includegraphics[width=1\linewidth]{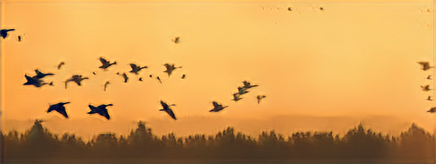}
    \end{subfigure} 

    \begin{subfigure}[t]{0.24\textwidth}
    \centering
    \includegraphics[width=1\linewidth]{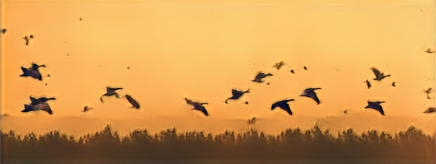}
    \end{subfigure}

    \begin{subfigure}[t]{0.12\textwidth}
    \centering
    \includegraphics[width=1\linewidth]{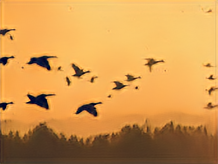}
    \end{subfigure}

    \\
        
    \begin{subfigure}[t]{0.12\textwidth}
    \centering
    \includegraphics[width=1\linewidth]{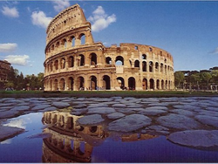}
    \end{subfigure}

    &

    \begin{subfigure}[t]{0.24\textwidth}
    \centering
    \includegraphics[width=1\linewidth]{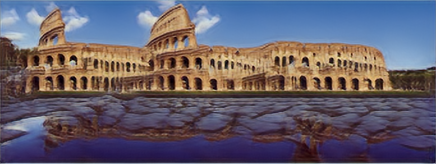}
    \end{subfigure}

    \begin{subfigure}[t]{0.12\textwidth}
    \centering
    \includegraphics[width=1\linewidth]{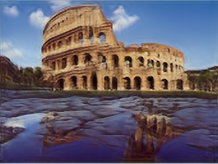}
    \end{subfigure}

    \begin{subfigure}[t]{0.24\textwidth}
    \centering
    \includegraphics[width=1\linewidth]{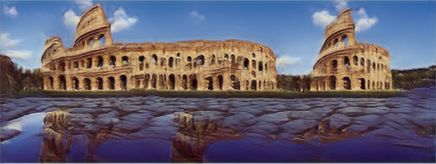}
    \end{subfigure}

    \begin{subfigure}[t]{0.12\textwidth}
    \centering
    \includegraphics[width=1\linewidth]{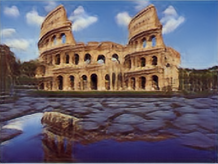}
    \end{subfigure}
    
    \\
        
    \begin{subfigure}[t]{0.12\textwidth}
    \centering
    \includegraphics[width=1\linewidth]{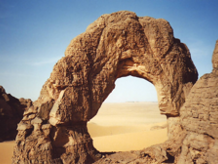}
    \end{subfigure}

    &

    \begin{subfigure}[t]{0.12\textwidth}
    \centering
    \includegraphics[width=1\linewidth]{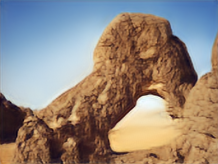}
    \end{subfigure}

    \begin{subfigure}[t]{0.24\textwidth}
    \centering
    \includegraphics[width=1\linewidth]{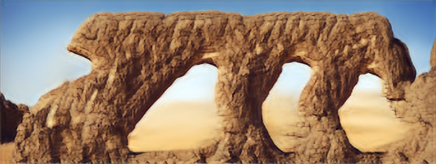}
    \end{subfigure}

    \begin{subfigure}[t]{0.12\textwidth}
    \centering
    \includegraphics[width=1\linewidth]{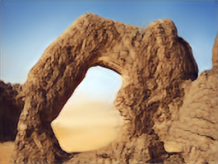}
    \end{subfigure}

    \begin{subfigure}[t]{0.24\textwidth}
    \centering
    \includegraphics[width=1\linewidth]{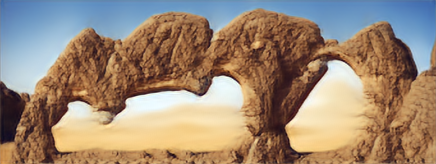}
    \end{subfigure}

    \end{tabular}
\caption{\textbf{BlendGAN trained on four images.} Our approach enables to train a single-image GAN model on multiple images, and then drawing random samples from each image identity separately.}
\label{fig:multi}
\end{figure*}{}

%% file: figures/clip/clip.tex
\begin{figure*}[ht]
\centering
\captionsetup[subfigure]{labelformat=empty,justification=centering,aboveskip=1pt,belowskip=1pt}
    \begin{tabular}[c]{c c c}

    \begin{subfigure}[t]{0.16\textwidth}
    \centering
    \caption*{Image \# 1}
    \end{subfigure}
    
    &

    \begin{subfigure}[t]{0.64\textwidth}
    \centering
    \caption*{Image morphing \textit{without} semantic blending constraint }
    \end{subfigure}
        
    &

    \begin{subfigure}[t]{0.16\textwidth}
    \centering
    \caption*{Image \# 2}
    \end{subfigure}

    \\
        
    \begin{subfigure}[t]{0.16\textwidth}
    \centering
    \includegraphics[width=1\linewidth]{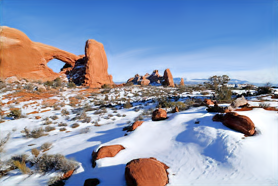}
    \end{subfigure}

    &

    \begin{subfigure}[t]{0.16\textwidth}
    \centering
    \includegraphics[width=1\linewidth]{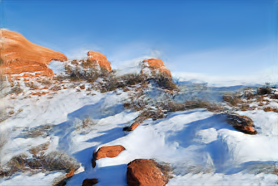}
    \end{subfigure}

    \begin{subfigure}[t]{0.16\textwidth}
    \centering
    \includegraphics[width=1\linewidth]{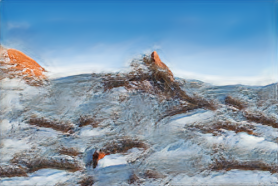}
    \end{subfigure}

    \begin{subfigure}[t]{0.16\textwidth}
    \centering
    \includegraphics[width=1\linewidth]{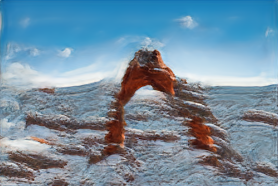}
    \end{subfigure}

    \begin{subfigure}[t]{0.16\textwidth}
    \centering
    \includegraphics[width=1\linewidth]{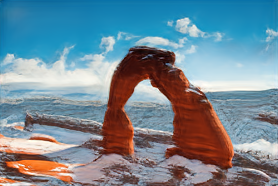}
    \end{subfigure}

    &

    \begin{subfigure}[t]{0.16\textwidth}
    \centering
    \includegraphics[width=1\linewidth]{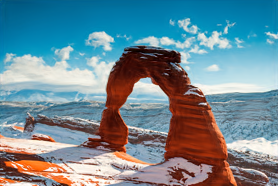}
    \end{subfigure}

    \\

    \vspace{-3.5mm}
        
    \begin{subfigure}[t]{0.16\textwidth}
    \centering
    \end{subfigure}

    &

    \begin{subfigure}[t]{0.16\textwidth}
    \centering
    \includegraphics[width=1\linewidth]{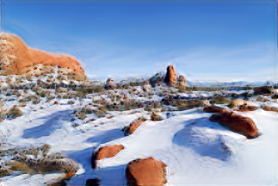}
    \end{subfigure}

    \begin{subfigure}[t]{0.16\textwidth}
    \centering
    \includegraphics[width=1\linewidth]{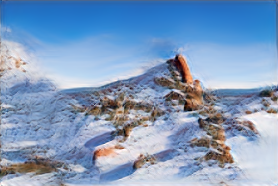}
    \end{subfigure}

    \begin{subfigure}[t]{0.16\textwidth}
    \centering
    \includegraphics[width=1\linewidth]{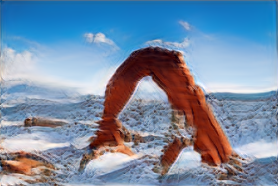}
    \end{subfigure}

    \begin{subfigure}[t]{0.16\textwidth}
    \centering
    \includegraphics[width=1\linewidth]{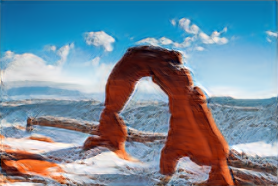}
    \end{subfigure}

    &

    \begin{subfigure}[t]{0.16\textwidth}
    \centering
    \end{subfigure}

    \\

    \vspace{-1.0mm}

    &

    \begin{subfigure}[t]{0.64\textwidth}
    \centering
    \caption*{Image morphing \textit{with} semantic blending constraint}
    \end{subfigure}
        
    &

    \end{tabular}
\caption{\textbf{Closing the gap between internal distributions.} When the training images contain distinct semantics, blending may result in artifacts (first row). This is because BlendGAN does not have an access to any additional semantics knowledge, and therefore struggles to close the gap between the two distributions. In such cases incorporating additional semantic loss improves the results (second row).}
\label{fig:clip}
\end{figure*}{}

%% file: sections/conclusion.tex
\section{Conclusion}

We introduced BlendGAN, a new framework to train a single-image GAN model on multiple images such that generation is conditioned on the image identity. Our model is able to generate realistic samples for each image separately, while having the same capacity as common single-image GANs models. Moreover, our model opens the door to applications that are not addressable by the original models. These correspond to mixing image identities across space, time, or scale, and can be applied to either random samples or the training images themselves.